\documentclass[letterpaper]{article} 
\usepackage{aaai25}  
\usepackage{times}  
\usepackage{helvet}  
\usepackage{courier}  
\usepackage[hyphens]{url}  
\usepackage{graphicx} 
\usepackage{natbib}  
\usepackage{caption} 
\usepackage{algorithm}
\usepackage{algorithmic}

\usepackage{newfloat}
\usepackage{listings}
\usepackage{enumitem}
\usepackage{amsmath}
\usepackage{multirow}
\usepackage{amssymb}
\usepackage{booktabs}
\usepackage{tabularx}

\DeclareCaptionStyle{ruled}{labelfont=normalfont,labelsep=colon,strut=off} 
\floatstyle{ruled}
\newfloat{listing}{tb}{lst}{}
\floatname{listing}{Listing}
\title{Thought-Path Contrastive Learning via Premise-Oriented Data Augmentation for Logical Reading Comprehension}
\author{
    Chenxu Wang\textsuperscript{\rm 1}, Ping Jian\thanks{Corresponding author.}\textsuperscript{\rm 1,2}, Zhen Yang\textsuperscript{\rm 1}
}
\affiliations{
    \textsuperscript{\rm 1}School of Computer Science and Technology, Beijing Institute of Technology, Beijing, China\\
    \textsuperscript{\rm 2}Beijing Engineering Research Center of High Volume Language Information Processing \\ and Cloud Computing Applications, Beijing Institute of Technology, Beijing, China \\
    \{wangchenxu, pjian, bityangzhen\}@bit.edu.cn

}

\usepackage{bibentry}

\begin{document}

\maketitle

\begin{abstract}
Logical reading comprehension is a challenging task that involves understanding the underlying semantics of text and applying reasoning to deduce the correct answer. Prior researches have primarily focused on enhancing logical reasoning capabilities through Chain-of-Thought (CoT) or data augmentation. However, previous work constructing chain-of-thought rationales concentrates solely on analyzing correct options, neglecting the incorrect alternatives. Addtionally, earlier efforts on data augmentation by altering contexts rely on rule-based methods, which result in generated contexts that lack diversity and coherence. To address these issues, we propose a Premise-Oriented Data Augmentation (PODA) framework. This framework can generate CoT rationales including analyses for both correct and incorrect options, while constructing diverse and high-quality counterfactual contexts from incorrect candidate options. We integrate summarizing premises and identifying premises for each option into rationales. Subsequently, we employ multi-step prompts with identified premises to construct counterfactual context. To facilitate the model's capabilities to better differentiate the reasoning process associated with each option, we introduce a novel thought-path contrastive learning method that compares reasoning paths between the original and counterfactual samples. Experimental results on three representative LLMs demonstrate that our method can improve the baselines substantially across two challenging logical reasoning benchmarks (ReClor and LogiQA 2.0).
\end{abstract}

\begin{links}
    \link{Code}{https://github.com/lalalamdbf/TPReasoner}
\end{links}

\section{Introduction}

\begin{figure}[t]
    \centering
    \includegraphics[width=\columnwidth]{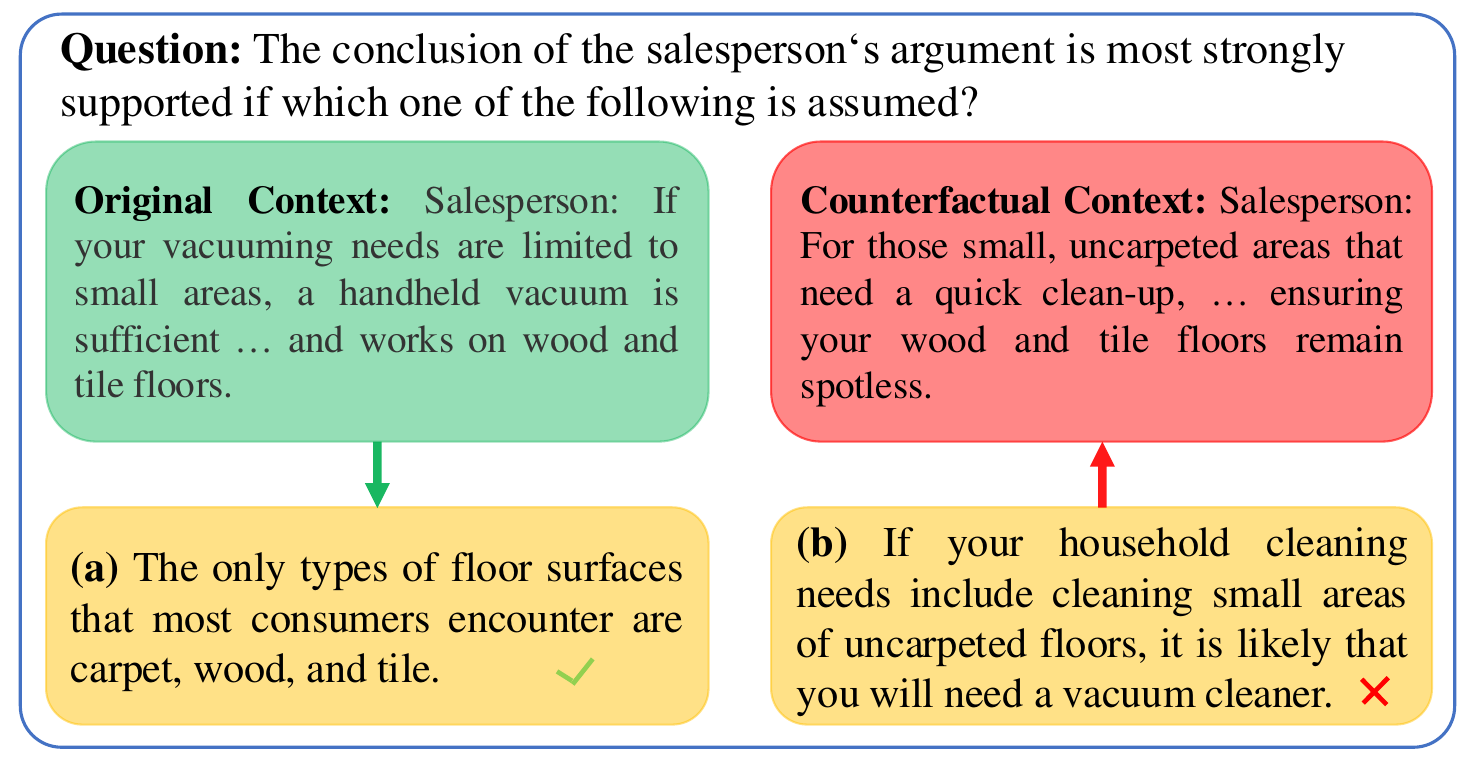}
    \caption{Generating counterfactual context from an incorrect candidate option.}
  \label{fig1:comparison} 
\end{figure}

Logical reasoning is a fundamental component of human cognition, essential for comprehending text and applying reasoning to deduce appropriate conclusions. Recently, challenging logical reasoning benchmarks have been proposed through machine reading comprehension (MRC) tasks \cite{yu2020reclor,liu2023logiqa}, which require models to derive the correct answer based on the given context, question and options. With the advent of large language models (LLMs), enhancing their capabilities in logical reasoning is a crucial step toward achieving strong artificial intelligence \cite{chollet2019measure}. Especially, the highly advanced model, GPT-4 \cite{achiam2023gpt}, has exhibited remarkable abilities to handle such tasks. However, a broad spectrum of open LLMs, including LLaMA2 \cite{touvron2023llama}, Mistral \cite{jiang2023mistral} and LLaMA3 \cite{llama3modelcard}, still fall short in logical reasoning, significantly trailing behind GPT-4. Consequently, improving the logical reasoning capabilities of community models has increasingly attracted the attention of many researchers \cite{liu2023logicot,jiang2023mistral}.

\begin{table}[t]
\centering
\setlength{\tabcolsep}{10.0pt}
\begin{tabular}{|p{0.90\linewidth}|}
\toprule
\textbf{Context:} [Content of the context]  \\ \hline
\textbf{Question:} [Content of the question]  \\ \hline
\textbf{Options:} [Content of the options]  \\ \hline
\textbf{Summarize Premises:} \\
1. [Premise 1] \\
2. [Premise 2] \\
3. [Premise 3] \\
\textbf{Analyze Options:} \\
(a) [Thought-path 1] \\
Identify Premises : Unrelated to the premises. \\
(b) [Thought-path 2] \\
Identify Premises: Supported by premises 2 and 3. \\
(c) [Thought-path 3] \\ 
Identify Premises: Unrelated to the premises. \\
(d) [Thought-path 4] \\
Identify Premises: Contradicted by premise 1. \\

[A summary of thought-paths]. Therefore, the optimal correct answer is (b). \\

\bottomrule
\end{tabular}

\caption{A logical reasoning example. The CoT rationale is annotated by GPT-3.5 or GPT-4. Due to space constraints, we refer to the specific reasoning process as [thought-path].}
\label{table:example}

\end{table}

For logical MRC tasks, LogiCoT \cite{liu2023logicot} constructs instruction-tuning data with Chain-of-Thought (CoT) rationales. Nevertheless, these rationales only provide analyses for the correct options, neglecting the incorrect alternatives. This oversight limits the model's ability to fully understand why certain answers are wrong, which is crucial for enhancing its reasoning capabilities and overall performance in distinguishing between similar options. In addition, previous studies typically create counterfactual contexts based on rule-based data augmentation. For instance, LReasoner \cite{wang-etal-2022-logic} generates logically nonequivalent sentences by utilizing templates and syntax parsing. AMR-LDA \cite{bao2023contrastive} constructs counterfactual sentences based on Abstract Meaning Representation (AMR, \citeauthor{banarescu2013abstract}, \citeyear{banarescu2013abstract}) graph and logical laws. These methods rely on complex principles and make minimal changes to the text, that cannot ensure the diversity of generated content and accurate modifications to its underlying logic. Additionally, they directly modify the context without considering its relationship with the options, which leads to a mismatch between the counterfactual context and options.

In view of above challenges, we propose a premises-oriented data augmentation (PODA) framework. As shown in Figure \ref{fig1:comparison} and Table \ref{table:example}, the objective of PODA is to generate CoT rationales that include analyses for both correct and incorrect options, while also constructing counterfactual contexts based on incorrect candidate options. In Table \ref{table:example}, analyses for both correct and incorrect options are presented in \textit{Analyze options}. Besides, we incorporate summarizing premises and identifying premises for each option into rationales. Each option has a specific relationship with these premises—either supported, contradicted, or unrelated. PODA will create high-quality and diverse counterfactual contexts using multi-step prompts based on these premises and relationships. Furthermore, since supervised fine-tuning (SFT) focuses solely on individual instances, it lacks the comparison between different samples. For original and counterfactual samples, there are thought-paths that indicate similar and dissimilar reasoning processes associated with options. Therefore, we propose a thought-path contrastive learning approach, which specifically compares thought-paths across different samples, facilitating the model's capabilities to better distinguish diverse reasoning paths. The main contributions of this paper are summarized as follows:

\begin{itemize}[leftmargin=*]
   \item   We propose a premise-oriented data augmentation framework, which can generate CoT rationales involving analyses for both correct and incorrect options, while automatically constructing diverse and high-quality counterfactual data from incorrect candidate options.
   \item We introduce a thought-path contrastive learning approach, facilitating models to distinguish different reasoning paths between original and counterfactual samples.
   \item Experimental results conducted on representative open LLMs (LLaMA2-7B, Mistral-7B and LLaMA3-8B) demonstrate that our method achieves superior performance on two logical MRC benchmarks.
 \end{itemize}

\section{Related Work}

\subsection{Chain-of-Thought Prompting}
LLMs are capable of performing complex reasoning to derive the final answer by generating intermediate reasoning steps through a process called Chain-of-Thought (CoT). Zero-shot-CoT \cite{kojima2022large} showcases impressive reasoning performance only using a single instruction "Let's think step by step". Few-shot-CoT \cite{zhang2022automatic,wang2023cue} further boosts the reasoning abilities of LLMs by incorporating several CoT demonstrations. In addition, by offering carefully-crafted CoT demonstrations, LLMs can be encouraged to develop the similar reasoning skills and deliver responses in a uniform format. To ensure obtained CoTs are well-structured, we adopt Few-shot-CoT for data collection using GPT-3.5 and GPT-4. Recently, \citet{liu-etal-2023-logicot} also utilized GPT-4 to annotate the intermediate steps of correct options for logical MRC tasks. In contrast, our study expands the analysis to include incorrect options and focuses on mining information from CoT rationales to generate new logical MRC data.

\subsection{Logical Reasoning}
Leveraging logical reasoning capabilities embodies a comprehensive approach to natural language understanding (NLU). Previous studies have primarily focused on integrating logical knowledge into language models. For example, \citet{huang-etal-2021-dagn} exploited a logic graph to model semantic relationships. \citet{wang-etal-2022-logic} and \citet{bao2023contrastive} constructed equivalent/nonequivalent instances through intricate logic rules and entity replacement. These techniques, however, are constrained by their reliance on manually designed rules, which struggles to reliably identify complex logical relationships in diverse texts. Thus, our work shifts away from annotating logical relationships. Instead, we decompose and construct contexts using premises as the foundational units. Moreover, our contrastive learning approach improves LLMs' logical reasoning capabilities by enabling them to distinguish various thought-paths.

\section{Methodology}

\begin{figure*}[t]

    \centering
    \includegraphics[width=2.1\columnwidth]{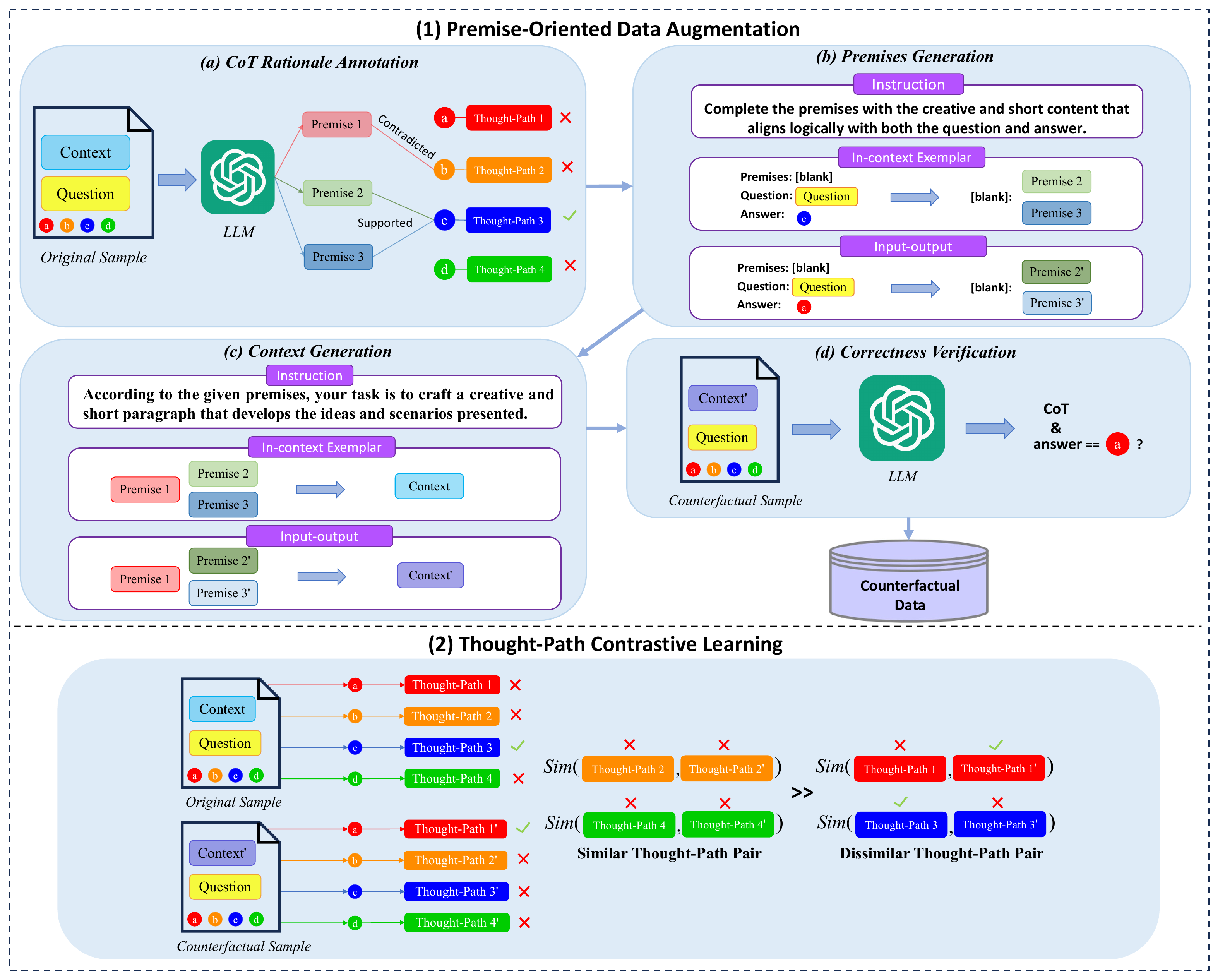}
    \caption{The overall architecture of our method. (1) PODA annotates Chain-of-Thought (CoT) rationales and generates counterfactual logical reasoning data. (2) The original and counterfactual samples are used for thought-path contrastive learning.}
    \label{fig3:overall}
\end{figure*}

Figure \ref{fig3:overall} shows the overall architecture of our method (PODA-TPCL). It consists of two key components: Premise-Oriented Data Augmentation (PODA) and Thought-Path Contrastive Learning (TPCL). The former module is aimed at generating CoT rationales that comprise analyses for correct and incorrect options, while constructing diverse and high-quality counterfactual logical reasoning data from incorrect candidate options. The latter one enhances the reasoning capabilities by comparing thought-paths between the original and counterfactual samples.

\subsection{Premise-Oriented Data Augmentation}

PODA initially creates analyses by forming thought-paths for both correct and incorrect options. Summarizing premises and identifying premises for each option are incorporated into CoT rationales, which are essential for generating new data. The core idea of it is to prompt a large language model through in-context learning to generate counterfactual data that can reverse the current answer to a new answer. A context can be divided into \textit{Premises} (the known information from the text), which have specific relationships with the options. The relationships are categorized into three types: \textit{supported}, \textit{contradicted} and \textit{unrelated}. Utilizing the premise as a foundational unit, we can construct counterfactual samples based on these relationships. 


\subsubsection{CoT Rationale Annotation}
As illustrated in Table \ref{table:example}, we design a structural CoT consisting of three steps: (1) \textit{Summarize Premises}: Extract supporting statements from the context to serve as premises. (2) \textit{Analyze Options}: Conduct a thorough evaluation of each option, clarifying the specific relationships between the options and premises, referred to as \textit{thought-path} in our work. (3) \textit{Derive answer}: Combine all thought-paths and determine the final answer. To guarantee a well-structured CoT, we utilize Few-shot-CoT for data collection.\footnote{In some cases where the model generates an incorrect response, we provide the correct answer to assist in its self-correction and refinement of the reasoning process.}

\subsubsection{Premises Generation}
To prompt GPT-4 for new premises generation, we use a masked natural language inference (NLI) format to build the prompt. Let $P_a$ represent the premises, associated with the question $Q$ and the current answer $a$. We replace $P_a$ with a mask token \textbf{[blank]}, and then $P_a$ serves as the output of the in-context exemplar to satisfy the question and answer. Given a new answer $a'$ we want to flip to, we ask the model to complete \textbf{[blank]} with creative premises $P_{a'}$ that align with $a'$. This approach enables the model to generate counterfactual premises that are logically consistent with the new answer.

\subsubsection{Context Generation}
To create new contexts, we preserve the origin premises $P_{\neg a}$ that are irrelevant to the current answer $a$, and introduce counterfactual premises $P_{a'}$ corresponding to the new answer $a'$. Let origin context $C$ serve as the output of the in-context exemplar, which is consistent with all origin premises $P = \{P_a, P_{\neg a}\}$. According to the reorganized premises $P' = \{P_{a'}, P_{\neg a}\}$, we ask the model to craft a creative context $C'$ that develops the ideas and scenarios presented. This newly crafted context integrates the counterfactual premises while maintaining coherence and expanding upon the original narrative structure.

\subsubsection{Correctness Verification}
\label{sub:cv}
Upon a combination of the initial three stages, we then implement correctness verification using Few-shot-CoT to filter out incorrect samples. The prompt and output format of this stage align with CoT Rationale Annotation. The potential mistakes of samples primarily stem from the following three aspects:

1. Some options are excessively absolute in their wording (e.g., using \textit{must} or \textit{can't}), which conflicts with the nature of the question, making them unsuitable as correct answers.
 
2. Several options in the original samples are deliberately designed as incorrect choices that violate common sense. Thus, it is inappropriate to create new contexts based on these options.
  
3. Given the complexity of logical reasoning, it is challenging to ensure that the generated premises align perfectly with the expected answers for particularly difficult samples.


  
%
%

\subsection{Thought-Path Contrastive Learning}
Supervised fine-tuning (SFT) can notably improve the model's performance. However, SFT only focuses on single instances, which results in its lack of the comparison between different samples. For logical MRC tasks, PODA annotates Chain-of-Thought (CoT) rationales, offering analyses for both correct and incorrect options, while also generating counterfactual samples.  It can be observed that thought-paths exhibit similar and dissimilar reasoning processes associated with options in original and counterfactual samples.

In light of such motivation, we propose a thought-path contrastive learning approach. As depicted in the part (2) of Figure \ref{fig3:overall}, the original/counterfactual sample has four thought-paths, with each corresponding to one option. Three thought-paths indicate that the corresponding options are incorrect, while one thought-path suggests it is correct. Therefore, for the original and counterfactual sample pair, the reasoning processes of thought-paths 2 and 2$'$ (thought-paths 4 and 4$'$) are analogous, whereas those of thought-paths 1 and 1$'$ (thought-paths 3 and 3$'$) are different. The goal of our method is to pull similar thought-paths closer while pushing different ones far apart. Simultaneously, we seek to enhance the model's capabilities to precisely distinguish between pairs of thought-paths (e.g., similarity(thought-paths 4, 4$'$) $\gg$ similarity(thought-paths 2, 2$'$)).
\begin{table*}[t]
\centering
\setlength{\tabcolsep}{11.0pt}
\begin{tabular}{l|cccc|cc}
\toprule
\multicolumn{1}{c|}{\multirow{2}{*}{\textbf{Model}}}  & 
\multicolumn{4}{c|}{\textbf{ReClor}}   & \multicolumn{2}{c}{\textbf{LogiQA 2.0}}   \\
     & Dev    & Test   & Test-E    & Test-H  & Dev  & Test      \\
\midrule
\multicolumn{7}{l}{\textbf{Discriminative Language Models}} \\
\, RoBERTa-Large \cite{liu2019roberta}   & 62.60 & 55.60 & 75.50    & 40.00  & -       & -                           \\
\, DGAN \cite{huang-etal-2021-dagn}    & 65.80          & 58.30              & 75.91      & 44.46  & -        & -                                   \\
\, LReasoner \cite{wang-etal-2022-logic}      & 66.20          & 62.40         & \textbf{81.40}      & 47.50 & -        & -                                     \\
\, AMR-LDA \cite{bao2023contrastive}        & 65.26         & 56.86              & 77.34    & 40.77  & -          & -    \\
\, FocalReasoner \cite{ouyang2024fact}        & 66.80       & 58.90            & 77.10    & 44.60  & -          & -    \\
\midrule   
\multicolumn{7}{l}{\textbf{Instruction-tuned LLMs}} \\                            
\, LLaMA2-7B-logicot \cite{liu2023logicot}      & 49.20          & 50.50         & 59.06      & 43.75        & 45.06            & 43.19   \\
\, LLaMA3-8B-logicot     & 61.50          & 62.65        & 71.25      & 55.89        & 54.11             & 54.07             \\
\, Mistral-7B-logicot      & 63.00          & 61.90         & 70.45      & 55.18       & 55.83             & 54.25              \\
\midrule
\multicolumn{7}{l}{\textbf{API-based LLMs (3-shot-CoT)}} \\

\, GPT-3.5 (gpt-3.5-turbo-0613)    & 56.00          & 58.20         & 61.82         & 55.36  &  55.07  &  51.15                        \\
\, GPT-4 (gpt-4o)   &  87.20        &  89.30             & 90.45         &  88.39  & 76.32   & 74.81 \\
\midrule
\multicolumn{7}{l}{\textbf{TPReasoner}} \\ 
\, LLaMA2-7B  & 58.00 & 58.63 & 66.14 & 52.74 & 49.71 & 49.75    \\
\, LLaMA3-8B  & \underline{67.60} & \underline{70.97} & 77.27 & \textbf{66.01} & \underline{60.29} & \underline{58.78}   \\
\, Mistral-7B  & \textbf{69.73} & \textbf{71.17} & \underline{78.79} & \underline{65.18} & \textbf{61.22} & \textbf{60.28}    \\
\bottomrule
\end{tabular}
\caption{Experimental results (Accuracy \%) of our method compared with baseline models on ReClor and LogiQA 2.0 benchmarks. Segment-1: Discriminative language models; Segment-2: Instruction-tuned LLMs; Segment-3: API-based LLMs (3-shot-CoT); Segment-4: TPReasoner (our method). Test-E and Test-H denote Test-Easy and Test-Hard respectively. The best and second best results are marked in bold and underlined (comparisons do not include API-based LLMs).}
\label{table:overall_comparison}

\end{table*}

Inspired by recent advances in learning to preference optimization algorithms such as RLHF \cite{ouyang2022training} and DPO \cite{rafailov2023direct}, our objective is to present a simple approach for comparing the similarity of thought-path pairs. To achieve this objective, we employ the Bradley-Terry preference model \cite{bradley1952rank} to construct the loss function for the similarity comparison. Given the input pair $\pi_0 = (x_1, x_2)$, the Bradley-Terry model calculates the likelihood of similarity comparison over thought-path pairs, denoted as $(p^{}_s, p'_{s}) > (p^{}_d, p'_{d}) \mid \pi_0$, where $(p^{}_s, p'_{s})$ and $(p^{}_d, p'_{d})$ represent the similar and different thought-path pairs, respectively. In our work, we simply choose the reward function \( r^{*} = \frac{1}{\tau}\text{sim}(\cdot) \) to measure similarity using cosine distance, where $\tau$ is the temperature coefficient controlling the sharpness of the similarity distribution. Under the Bradley-Terry model, we can derive a streamlined probability measure for pairwise similarity comparison:

\begin{equation}
\begin{aligned}
&p^*((p^{}_s,  p'_{s})  > (p^{}_d, p'_{d}) \mid \pi_0) \\ &=\sigma (r(p^{}_s, p'_{s}) - r(p^{}_d, p'_{d})) \\
&= \frac{1}{1 + \exp [{\frac{1}{\tau}}\text{sim}(p^{}_d,  p'_{d}) - \frac{1}{\tau}\text{sim}(p^{}_s, p'_{s})]} 
\label{eq:1}
\end{aligned}
\end{equation}

where $\pi_0$ is omitted as it doesn't directly contribute to the calculation. Then we formulate the problem as binary classification using the negative log-likelihood loss:

\begin{equation}
\begin{aligned}
  &\mathcal{L}((p^{}_s,  p'_{s}), (p^{}_d,  p'_{d}), \pi_0)) \\
  &= -\mathbb{E}[\log(\sigma(r(p^{}_s,  p'_{s})-r(p^{}_d,  p'_{d})))]
  \label{eq:2}
\end{aligned}
\end{equation}

The objective of this loss function is to decrease the distance between dissimilar pair $(p^{}_d,  p'_{d})$ while increasing the distance between similar pair $(p^{}_s,  p'_{s})$. Additionally, the model learns to differentiate between pairs of thought-pairs based on preference optimization. Due to the presence of multiple groups of similar and dissimilar thought-path pairs for input $\pi_0$, we simply calculate the average loss as follows:

\begin{equation}
\begin{aligned}
  &\mathcal{L}_\text{TPCL}((p^{}_s,  p'_{s}), (p^{}_d,  p'_{d}), \pi_0)) = \\
   &-\mathbb{E}[\frac{1}{N*M}\sum\limits_{i=1}^N\sum\limits_{j=1}^M\log(\sigma(r(p^{}_{sj},  p'_{sj}) \\&- r(p^{}_{di},  p'_{di})))]
  \label{eq:3}
\end{aligned}
\end{equation}

where $N$ and $M$ represent the number of similar and dissimilar thought-path pairs respectively. Both $N$ and $M$ are set to 2 in our work. We also add a cross-entropy loss consistent with SFT to ensure the model does not deviate from the data distribution. Given an input sequence $x$, the average likelihood of generating the output sequence $y$, consisting of $m$ tokens, is computed as follows:

\begin{equation}
\begin{aligned}
  \mathcal{L}&_\text{SFT} = \frac{1}{m} \sum_{t=1}^{m} \log P (y_t \mid x, y_{<t})
  \label{eq:4}
\end{aligned}
\end{equation}

The overall training goal is the combination of TPCL loss and SFT loss:

\begin{equation}
\begin{aligned}
  \mathcal{L} = \mathcal{L}&_\text{TPCL} + \mathcal{L}_\text{SFT}
  \label{eq:5}
\end{aligned}
\end{equation}

\begin{table*}[t]
\centering
\setlength{\tabcolsep}{16pt}
\begin{tabular}{l|cccc|cc}
\toprule
\multicolumn{1}{c|}{\multirow{2}{*}{\textbf{Model}}}  & 
\multicolumn{4}{c|}{\textbf{ReClor}}   & \multicolumn{2}{c}{\textbf{LogiQA 2.0}}   \\
     & Dev    & Test   & Test-E    & Test-H  & Dev  & Test      \\
\midrule
LLaMA2-7B  & \textbf{58.00} & \textbf{58.63} & \textbf{66.14} & \textbf{52.74} & \textbf{49.71} & \textbf{49.75}    \\
- w/o TPCL  & 55.27 & 56.73 & 63.48 & 51.43 & 48.25 & 48.60   \\ 
- w/o TPCL + CD   & 53.20 & 52.17 & 59.16 & 46.67 & 46.15 & 45.48   \\
- w/o TPCL + CD + WOA   & 51.40 & 50.27 & 56.21 & 45.58 & 44.89 & 44.47   \\ \hline

LLaMA3-8B  & \textbf{67.60} & \textbf{70.97} & \textbf{77.27} & \textbf{66.01} & \textbf{60.29} & \textbf{58.78}   \\
- w/o TPCL  & 66.00 & 69.30 & 74.09 & 65.54 & 57.11 & 57.06    \\ 
- w/o TPCL + CD   & 63.07 & 66.10 & 73.11 & 60.59 & 55.19 & 54.83\\
- w/o TPCL + CD + WOA   & 61.17 & 64.03 & 71.53 & 58.11 & 53.51 & 53.07 \\

\hline
Mistral-7B  & \textbf{69.73} & \textbf{71.17} & \textbf{78.79} & \textbf{65.18} & \textbf{61.22} & \textbf{60.28}    \\
- w/o TPCL  & 67.07 & 69.57 & 78.03 & 64.40 & 59.91 & 58.63    \\ 
- w/o TPCL + CD    & 65.20 & 67.37 & 75.68 & 62.92 & 58.24 & 56.97  \\
- w/o TPCL + CD + WOA    & 63.10 & 64.83 & 73.02 & 58.37 & 56.04 & 55.19  \\ 
\bottomrule
\end{tabular}
\caption{Ablation study of our method. TPCL stands for thought-path contrastive learning approach. CD refers to the utilization of counterfactual data. WOA signifies that CoT rationales involve the analyses for wrong options.}
\label{table:ablation_study}

\end{table*}

\section{Experiment}

\subsection{Datasets}
\textbf{ReClor} \cite{yu2020reclor} comprises 6,138 question-answering samples collected from standardized exams including GMAT and LSAT, which are split into train / dev / test sets with 4,638 / 500 / 1,000 samples respectively. To evaluate the difficulty of the questions, the test set is further divided into Test-E and Test-H. The instances on Test-E are easy and biased that can be solved without knowing contexts and questions. The other harder and unbiased ones are taken as the Test-H set.

\noindent\textbf{LogiQA 2.0} \cite{liu2023logiqa} is an updated and re-annotated version of LogiQA \cite{liu2020logiqa}. There are 15,708 instances derived from the Chinese Civil Service Examination, meticulously translated into English by experts. The dataset is randomly split into train / dev / test sets with 12,567 / 1,569 / 1,572 samples respectively.

\noindent\textbf{Synthetic Data} is generated by our PODA framework. We construct 5,075 and 13,477 counterfactual samples based on the train sets of ReClor and LogiQA 2.0, respectively. To perform thought-path contrastive learning, each counterfactual sample is paired with its corresponding origin one.

\subsection{Implementation Settings}
In PODA framework, gpt-3.5-turbo-0613 and gpt-4-0613 are utilized for CoT Rationale Annotation. Subsequently, gpt-4-0125-preview is employed for Premises and Context Generation. In the end, gpt-4-0613 is used for Correctness Verification.\footnote{We found that gpt-4-0613 exhibits a superior capability for generating data in a structural format compared to gpt-4-0125-preview. Hence, we chose gpt-4-0613 to annotate CoT rationales.} We set the sampling temperature of 0.75 and the top probability of 0.9, ensuring the generated text maintains both diversity and high quality. The detailed prompts are provided in Appendix C.2.

During the training process, we adopt LLaMA2-7B, Mistral-7B and LLaMA3-8B as baselines. In order to accelerate training, we employ LoRA \cite{hu2021lora} to fine-tune the model. The AdamW optimizer \cite{loshchilov2017decoupled} is used with a learning rate warmup of 0.03. Due to the absence of corresponding counterfactual samples for some original samples as illustrated in Correctness Verification, we implement a two-stage training strategy. The implementation of our code refers to Llamafactory \cite{zheng2024llamafactory}. All hyper-parameters of training are listed in Appendix A.
\subsection{Baselines}
In this paper, we compare our method with two types of baselines: discriminative language models and large language models (LLMs). The discriminative language model baselines include RoBERTa-Large \cite{liu2019roberta}, DGAN \cite{huang-etal-2021-dagn}, LReasoner \cite{wang-etal-2022-logic}, AMR-LDA \cite{bao2023contrastive} and FocalReasoner \cite{ouyang2024fact}. The LLM baselines encompass instruction-tuned method such as LogiCoT \cite{liu2023logicot}, as well as API-based LLMs like GPT-3.5 (gpt-3.5-turbo-0613) and GPT-4 (gpt-4o) by utilizing 3-shot-CoT. More details can be found in Appendix B.

\section{Result and Analysis}

\subsection{Overall Results}
Table \ref{table:overall_comparison} presents the primary experimental results of our method and other baselines on ReClor and LogiQA 2.0 benchmarks, in terms of accuracy. Our method employs the CoT rationales, which is appropriate for generative LLMs. Hence, we did not conduct experiments on discriminative language models. Compared to these baselines, TPReasoner exhibits superior performance except for GPT-4.

On ReClor dataset, LLaMA3-8B and Mistral-7B, based on our approach, significantly outperform all discriminative model methods. Compared with LReasoner, PODA-TPCL achieves improvements of 1.4-3.5\% and 8-9\% on the dev and test sets, respectively. Although LLaMA2-7B, when using our method, does not surpass all discriminative model baselines, the results on Test-H demonstrate that it exhibits stronger robustness and generalization for data distribution. There is a substantial disparity (exceeding 30\%) between the performances on Test-E and Test-H for discriminative models, indicating that these models tend to take shortcuts for simpler and biased samples rather than genuinely comprehending them. In contrast, our method achieves a gap of less than 15\% between Test-E and Test-H, with the performance on Test-H clearly surpassing that of the discriminative models. This demonstrates the great potential of leveraging CoT rationales to solve complex logical reasoning tasks. 

Compared to the instruction-tuned LLMs based on LogiCoT, our models achieve superior performance on ReClor and LogiQA 2.0 datasets. This improvement is attributed to the PODA framework, which offers additional analyses of incorrect options and constructs high-quality, diverse counterfactual instances. Additionally, TPCL boosts the model's reasoning capabilities by facilitating the learning of similar and distinct thought-paths between different samples. Furthermore, our model demonstrates competitive performance comparable to GPT-3.5, trailing only behind GPT-4.

\subsection{Ablation Study}
An ablation study is conducted to investigate the efficacy of three key components, thought-path contrastive learning (TPCL), counterfactual data (CD) and wrong options analyses (WOA), as presented in Table \ref{table:ablation_study}. For w/o TPCL, we eliminate TPCL and only employ SFT to train the model. There is a noticeable decline in performance, with a drop of 1-3\% across the two datasets. These results convincingly demonstrate that TPCL significantly boosts the model's reasoning capabilities by comparing reasoning paths between original and counterfactual samples. For w/o TPCL + CD, we additionally exclude the counterfactual samples generated by PODA and solely utilize the original data. It can be observed that the models without CD have severe performance degradation. This suggests that the counterfactual samples are beneficial for LLMs to conduct logical reasoning. Furthermore, it demonstrates that the data synthesized by our framework is of high-quality, automatically generated without requiring human interventions. For w/o TPCL + CD + WOA, we further omit the analyses of wrong options in CoT rationales. As a result, the models' performance decreases by approximately 2\%. It indicates that incorporating reasoning processes for incorrect options enables the model to analyze problems more thoroughly, thereby improving its reasoning abilities. Overall, PODA-TPCL achieves a performance improvement of 5-7\% across three models on ReClor and LogiQA 2.0 datasets, underscoring its exceptional robustness and generalization capabilities.

\subsection{Evaluation of Data Quality}

\begin{table}[t]
\centering
\setlength{\tabcolsep}{10.0pt}
\begin{tabular}{l|c}
\toprule
\textbf{Model} & \textbf{Accuracy}  \\ \midrule
Mixtral-8$\times$7B-Instruct & 74.05 \\
GPT-3.5 (gpt-3.5-turo-0613) & 75.50 \\
LLaMA2-70B-chat & 79.39 \\
LLaMA3-70B-Instruct & 82.19 \\
GPT-4 (gpt-4o-2024-05-13) & 93.00 \\ \midrule
Human Performance & 90.00 \\

\bottomrule
\end{tabular}

\caption{Evaluation of accuracy (\%) for counterfactual data.}
\label{table:evaluation_cd}

\end{table}

\subsubsection{Accuracy Evaluation of Counterfactual Data}
A primary concern is whether the synthetic data accurately matches the correctly labeled option. In order to evaluate this, we choose five outstanding LLMs, including Mixtral-8$\times$7B-Instruct, GPT-3.5 (gpt-3.5-turo-0613), LLaMA2-70B-chat, LLaMA3-70B-Instruct and GPT-4 (gpt-4o-2024-05-13). Non-GPT series models evaluate all counterfactual data. Due to budget constraints, a random subset of 200 samples from the generated dataset are evaluated by GPT-3.5 and GPT-4. We utilize 3-shot CoTs to evaluate the accuracy. As illustrated in Table \ref{table:evaluation_cd}, the accuracy assessed by LLaMA3-70B-Instruct and GPT-4 can reach 82.19\% and 93\% respectively, indicating PODA can generate high-quality counterfactual data. Moreover, accuracies for other models vary between approximately 75\% and 79\%, reflecting the significant challenges and complexities presented by these data. Overall, the generated data can have applicability in both training and evaluation domains.

\subsubsection{Comparison for Counterfactual Data} We compare PODA with a rule-based method, LReasoner \cite{wang-etal-2022-logic}, which constructs counterfactual contexts by modifying logical expressions according to logical laws. Two random subsets of 200 counterfactual samples were selected respectively. We ensure that the two subsets are derived from the same set of original samples for a fair comparison. These instances are evaluated by GPT-4 (gpt-4o-2024-05-13) using four key metrics: Coherence (Is the context well-connected and logically consistent?), Clarity (Is the context clear and easy to understand?), Relevance (Does the context relate to the question and options?), and Diversity (How does the counterfactual context differ from the original one?). Each context was rated on a scale from 1 (poor) to 5 (excellent) for each metric. Our method attains average scores of 4.61 for Coherence, 4.36 for Clarity, 4.71 for Relevance, 3.18 for Diversity, and 4.22 Overall. In contrast, LReasoner achieves average scores of 2.98 for Coherence, 2.96 for Clarity, 4.63 for Relevance, 1.08 for Diversity, and 2.91 Overall. This comparison clearly demonstrates that the contexts generated by our method significantly outperform those produced by the rule-based method in both quality and diversity.

\subsubsection{Comparison for CoT Rationales} We compare PODA with LogiCoT, which also utilizes GPT-4 for CoT rationales but focuses solely on analyzing the correct option. Similarly, two random subsets of CoT rationales were selected from the same contexts. These rationales are assessed by GPT-4 (gpt-4o-2024-05-13) using four key metrics: Coherence (Is the CoT rationale logically consistent?), Completeness (Does it offer a thorough explanation for the reasoning?), Relevance (Does it directly and effectively addresses the context, question and options?), and Faithfulness (Is it factually correct and free from fabricated details?). Each rationale was rated on a scale from 1 (poor) to 5 (excellent) for each metric. PODA surpasses LogiCoT across four metrics, especially in Completeness. This suggests that incorporating the analysis of incorrect options into the rationales can enhance the quality of CoTs.

\begin{table}[t]
\centering
\begin{tabular}{l|c|c|c|c|c}
\toprule
\textbf{Method} & \textbf{Coh} & \textbf{Clar} & \textbf{Rel} & \textbf{Div} & \textbf{Overall}  \\ \midrule
LReasoner & 2.98 & 2.96 & 4.63 & 1.08 & 2.91 \\
PODA & \textbf{4.61} & \textbf{4.36} & \textbf{4.71} & \textbf{3.18} & \textbf{4.22} \\

\bottomrule
\end{tabular}

\caption{Evaluating counterfactual data on Coherence (Coh), Clarity (Clar), Relevance (Rel), and Diversity (Div).}
\label{table:comparison_cd}

\end{table}

\begin{table}[t]
\centering
\begin{tabular}{l|c|c|c|c|c}
\toprule
\textbf{Method} & \textbf{Coh} & \textbf{Comp} & \textbf{Rel} & \textbf{Faith} & \textbf{Overall}  \\ \midrule
LogiCoT & 4.82 & 3.38 & 4.91 & 4.80 & 4.48 \\
PODA & \textbf{4.86} & \textbf{4.46} & \textbf{4.99} & \textbf{4.86} & \textbf{4.79} \\

\bottomrule
\end{tabular}

\caption{Evaluating rationales on Coherence (Coh), Completeness (Comp), Relevance (Rel), and Faithfulness (Faith).}
\label{table:comparison_cot}

\end{table}

\subsection{What Does TPCL Update Do?}
We analyze the gradient of the loss function $\mathcal{L}_\text{TPCL}$ (considering a group of similar and dissimilar thought-path pairs), whose gradient can be written as:

\begin{equation}
\begin{aligned}
  \nabla &\mathcal{L}_\text{TPCL}((p^{}_s,  p'_{s}), (p^{}_d,  p'_{d}), \pi_0)) = \\
  & -\frac{1}{\tau}\mathbb{E}[\sigma(r(p^{}_d,  p'_{d})-r(p^{}_s,  p'_{s}))\\ &\ast \, [\nabla \text{sim}(p^{}_s,  p'_{s}) - \nabla \text{sim}(p^{}_d,  p'_{d})]]
  \label{eq:6}
\end{aligned}
\end{equation}

Intuitively, the gradient of $\mathcal{L}_\text{TPCL}$ increases the similarity of the similar thought-paths $((p^{}_s,  p'_{s})$ and decreases the similarity of the dissimilar thought-paths $((p^{}_d,  p'_{d})$. Meanwhile, $\sigma(r(p^{}_d,  p'_{d})-r(p^{}_s,  p'_{s}))$ serves as an adjustable weight for the similarity reward estimate, assigning greater weight when $r(p_d, p'_d)$ is approximately equal to or greater than $r(p_s, p'_s)$.  This mechanism accelerates the convergence of the loss function. Overall, the gradient update aligns with our objective to pull similar thought-paths closer while pushing dissimilar ones further apart, which enhances the model's reasoning capabilities by comparing thought-paths between the original and counterfactual samples.
\section{Conclusion}

In this paper, we propose a premise-oriented data augmentation framework that generates CoT rationales, providing analyses for both correct and incorrect options. Additionally, the framework automatically constructs diverse and high-quality counterfactual data from incorrect candidate options. We also introduce a thought-path contrastive learning method to effectively leverage pairs of original and counterfactual samples, enabling models to distinguish between different reasoning paths. Extensive experiments conducted on three open LLMs demonstrate that our approach achieves competitive performance on two logical reasoning benchmarks.

\section*{Acknowledgments}

This work is supported by the grants from the National Natural Science Foundation of China (No. 62172044 and No. 62376130). The authors would like to thank the organizers of AAAI 2025 and the reviewers for their helpful suggestions.

\bibliography{aaai25}

\newpage
\appendix
\section{Hyper-parameters of Training}
During the training process, we set the batch size to 32, which is implemented using gradient accumulation. The maximum sequence length is truncated at 1536. In order to accelerate training, we employ LoRA to fine-tune the model. The LoRA matrices are appended in both the attention layer and the mlp layer. The dropout rate is set to 0.05 between the two matrices. The rank of LoRA is set to 64, with the alpha is set to 64 for ReClor and 32 for LogiQA 2.0. In the first stage, we only use supervised fine-tuning (SFT) to train the model with 2 epochs on all samples. The learning rate is set to 2e-4 for ReClor and 1e-4 for LogiQA 2.0. In the second stage, we incorporate TPCL, training the model for 1 epoch on sample pairs with a learning rate of 1e-6. The temperature coefficient $\tau$ is set to 0.1. We use the AdamW optimizer with a learning rate warmup of 0.03. The training and evaluation template is basically from Alpaca. The temperature is set as 0 to evaluate fine-tuned models. All experiments are conducted on one 80GB NVIDIA A800 GPU using three different random seeds, and we report the average values. The implementation of our code refers to Llamafactory.

\section{Baselines}
\textbf{(1) RoBERTa-Large:} The language model RoBERTa is fine-tuned on domain-specific datasets to obtain the predictions. 

\noindent\textbf{(2) DGAN:} It's a graph network to tackle logical reasoning tasks by encoding discourse information and learning discourse-aware features.

\noindent\textbf{(3) LReasoner:} It proposes a logic-driven context extension framework and a data augmentation algorithm based on extracted logical expressions.

\noindent\textbf{(4) AMR-LDA:} It proposes a logic-driven data augmentation approach, that constructs equivalent and nonequivalent sentences based on Abstract Meaning Representation (AMR) graph and logical laws. 

\noindent\textbf{(5) FocalReasoner:} It reveals the logical structures within the context and captures the interactions between the context and options to create a comprehensive supergraph for reasoning.

\noindent\textbf{(6) LLaMA2-7B-logicot:} It's a LLaMA2-7B model fine-tuned on logical MRC instances within LogiCoT, whose CoT rationales are collected by GPT-4 and focus on the reasoning for the correct options. For a fair comparison, we ensured that the training configuration aligns consistently with our method. Similarly, the Mistral-7B-logicot and LLaMA3-8B-logicot models have been fine-tuned using the same configuration.

\noindent\textbf{(7) GPT-3.5:} It is an exceptional model developed by OpenAI, designated as gpt-3.5-turbo-0613. We evaluate its logical reasoning capabilities utilizing 3-shot CoTs.

\noindent\textbf{(8) GPT-4:} It is the most advanced model created by OpenAI, designated as gpt-4o. We still evaluate its logical reasoning capabilities utilizing 3-shot CoTs.

\section{PODA}

\subsection{Workflow}
The premises-oriented data augmentation framework (PODA) consists of four stages: CoT Rationale Annotation, Premises Generation, Context Generation, and Correctness Verification. We illustrate the workflow of PODA in Algorithm \ref{alg:PODA}.

\subsection{Prompts for Generating Data}
The prompts used for CoT Rationale Annotation, Premises Generation, Context Generation, and Correctness Verification are detailed in Figures \ref{fig:cra}, \ref{fig:pg}, \ref{fig:cg} and \ref{fig:cv}, respectively. We configured the sampling temperature at 0.75 and set the top probability at 0.9.

\subsection{Prompts for Evaluating Data Quality}
The prompts for evaluating the quality of the counterfactual context cover Coherence (Is the context well-connected and logically consistent?), Clarity (Is the context clear and easy to understand?), Relevance (Does the context relate to the question and options?) and Diversity (How does the counterfactual context differ from the original one?). These prompts are depicted in Figures \ref{fig:context_co}, \ref{fig:context_cl}, \ref{fig:context_re} and \ref{fig:context_div}. Additionally, the prompts for evaluating the CoT rationales include assessments of Coherence (Is the CoT rationale logically consistent?), Completeness (Does it offer a thorough explanation for the reasoning?), Relevance (Does it directly and effectively addresses the context, question and options?) and Faithfulness (Is it factually correct and free from fabricated details?). These are shown in Figures \ref{fig:cot_coherence}, \ref{fig:cot_completeness}, \ref{fig:cot_relevance} and \ref{fig:cot_faithfulness}. We still set a sampling temperature of 0.75 and a top probability of 0.9.

\begin{algorithm}[htb]
\caption{Workflow of PODA}
\label{alg:PODA}
\begin{algorithmic}[1] 
\REQUIRE Samples $x^{1:N}$, CoT Rationale Annotation $CRA$,
        Premises Generation $PG$, Context Generation $CG$, Correctness Verification $CV$, $x$ = (Context $C$, Question $Q$, Options $O$, Current Answer $a$, New Answers $a'^{1:M}$)
\ENSURE Construct counterfactual data $CD$
\FOR{$i \leftarrow 1$ \TO $N$}
    \STATE $(C, Q, O, a, a'^{1:M}) \leftarrow x^i$
    \STATE $(P_a, P_{\neg a}) \leftarrow CRA(C, Q, O)$ $\triangleright$ $P_a$ and $P_{\neg a}$ denote
    whether the premises are related to $a$.
    \FOR{$j \leftarrow 1$ \TO $M$}
    \STATE $P_{a'} \leftarrow PG((Q, a, P_{a}), Q, a'^{j})$ $\triangleright$ $P_{a'}$ denotes the
    
    new premises aligning with $a'^{j}$.
    \STATE $C' \leftarrow CG((P_a, P_{\neg a}, C), P_{a'}, P_{\neg a})$ $\triangleright$ $C'$ denotes
    
    the new context corresponding to $a'^{j}$.
    \IF {$CV(C', Q, O) == a'^{j}$}
    \STATE $CD \leftarrow CD$.Append$(C', Q, O, a'^{j})$ 
    \ENDIF
    \ENDFOR
\ENDFOR
\STATE \textbf{return} $CD$
\end{algorithmic}
\end{algorithm}

\subsection{Human Evaluation}

We supplement the human evaluation of data quality, with all evaluators being undergraduate or master’s students.

\begin{table}[htb]
\centering
\begin{tabular}{l|c|c|c|c|c}
\toprule
\textbf{Method} & \textbf{Coh} & \textbf{Clar} & \textbf{Rel} & \textbf{Div} & \textbf{Overall}  \\ \midrule
LReasoner & 2.44 & 3.17 & 4.37 & 1.07 & 2.76 \\
PODA & \textbf{4.06} & \textbf{4.39} & \textbf{4.44} & \textbf{3.09} & \textbf{4.00} \\

\bottomrule
\end{tabular}

\caption{Human evaluation of counterfactual data on Coherence (Coh), Clarity (Clar), Relevance (Rel), and Diversity (Div).}
\label{table:human_comparison_cd}

\end{table}

\begin{table}[htb]
\centering
\begin{tabular}{l|c|c|c|c|c}
\toprule
\textbf{Method} & \textbf{Coh} & \textbf{Comp} & \textbf{Rel} & \textbf{Faith} & \textbf{Overall}  \\ \midrule
LogiCoT & 4.37 & 2.93 & 4.14 & 4.73 & 4.04 \\
PODA & \textbf{4.41} & \textbf{4.47} & \textbf{4.64} & \textbf{4.83} & \textbf{4.59} \\

\bottomrule
\end{tabular}

\caption{Human evaluation of rationales on Coherence (Coh), Completeness (Comp), Relevance (Rel), and Faithfulness (Faith).}
\label{table:human_comparison_cot}

\end{table}

\setcounter{figure}{0}

\begin{figure}[t]

    \centering
    \includegraphics[width=\columnwidth]{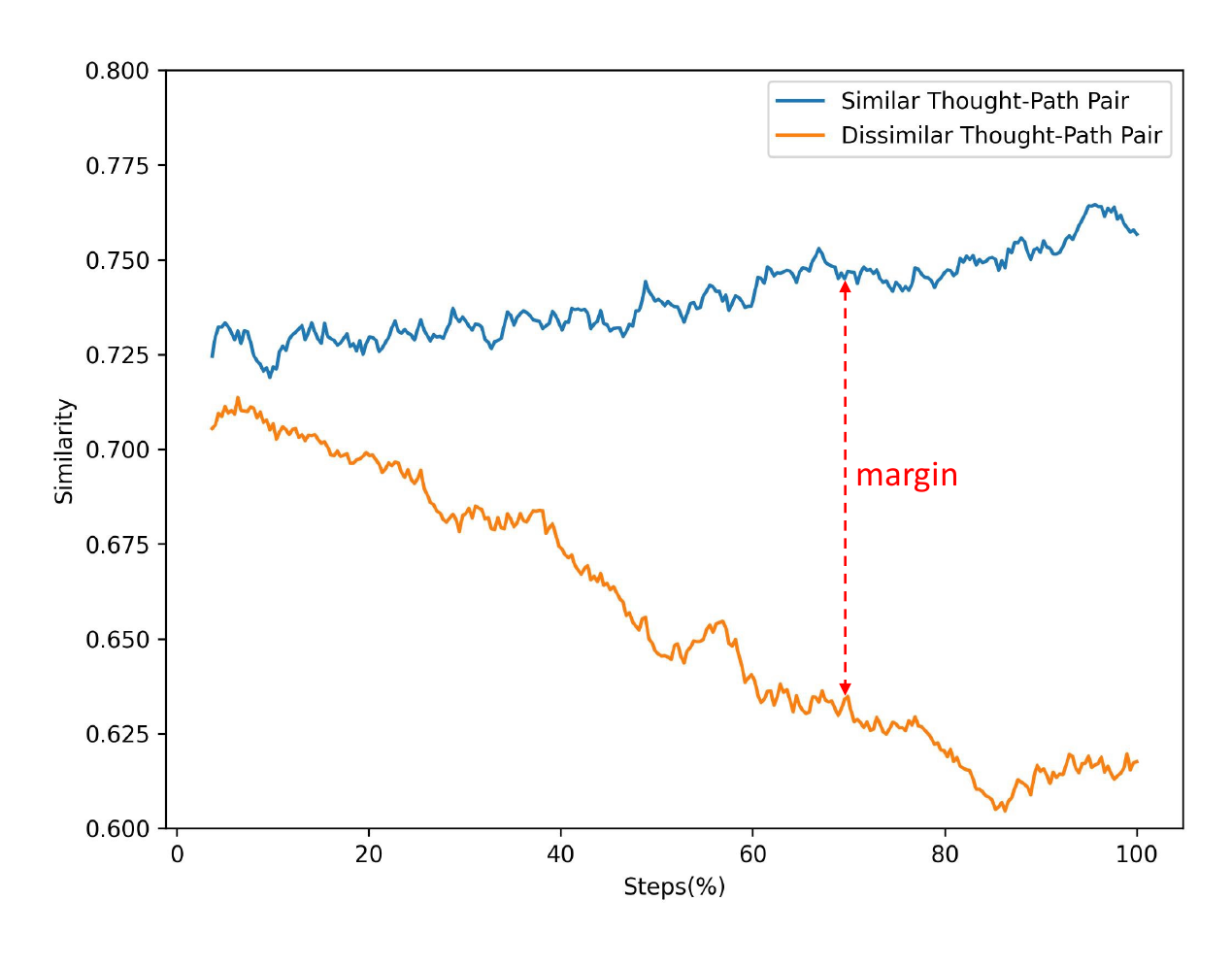}
    \caption{The trend of similarity variation for both similar and dissimilar thought-path pairs in TPCL.}
    \label{fig:trend}
\end{figure}

\section{Visualization of Similarity in TPCL}

Figure \ref{fig:trend} depicts the trend of similarity variation for both similar and dissimilar thought-path pairs. Initially, the margin between two types of thought-path pairs is minimal, indicating that the model struggles to effectively distinguish reasoning paths in its vector representation. With the training through TPCL, the similarity of similar thought-path pairs progressively increases, whereas the similarity between pairs of dissimilar thought-paths declines rapidly. This phenomenon aligns with the objective of TPCL, which aims to draw similar thought-paths closer together while distancing divergent ones. The margin between thought-pairs maintains a continuous increase, demonstrating the model's growing proficiency in distinguishing similar and dissimilar reasoning paths.

\begin{figure*}[htb]

    \centering
    \includegraphics[width=2.1\columnwidth]{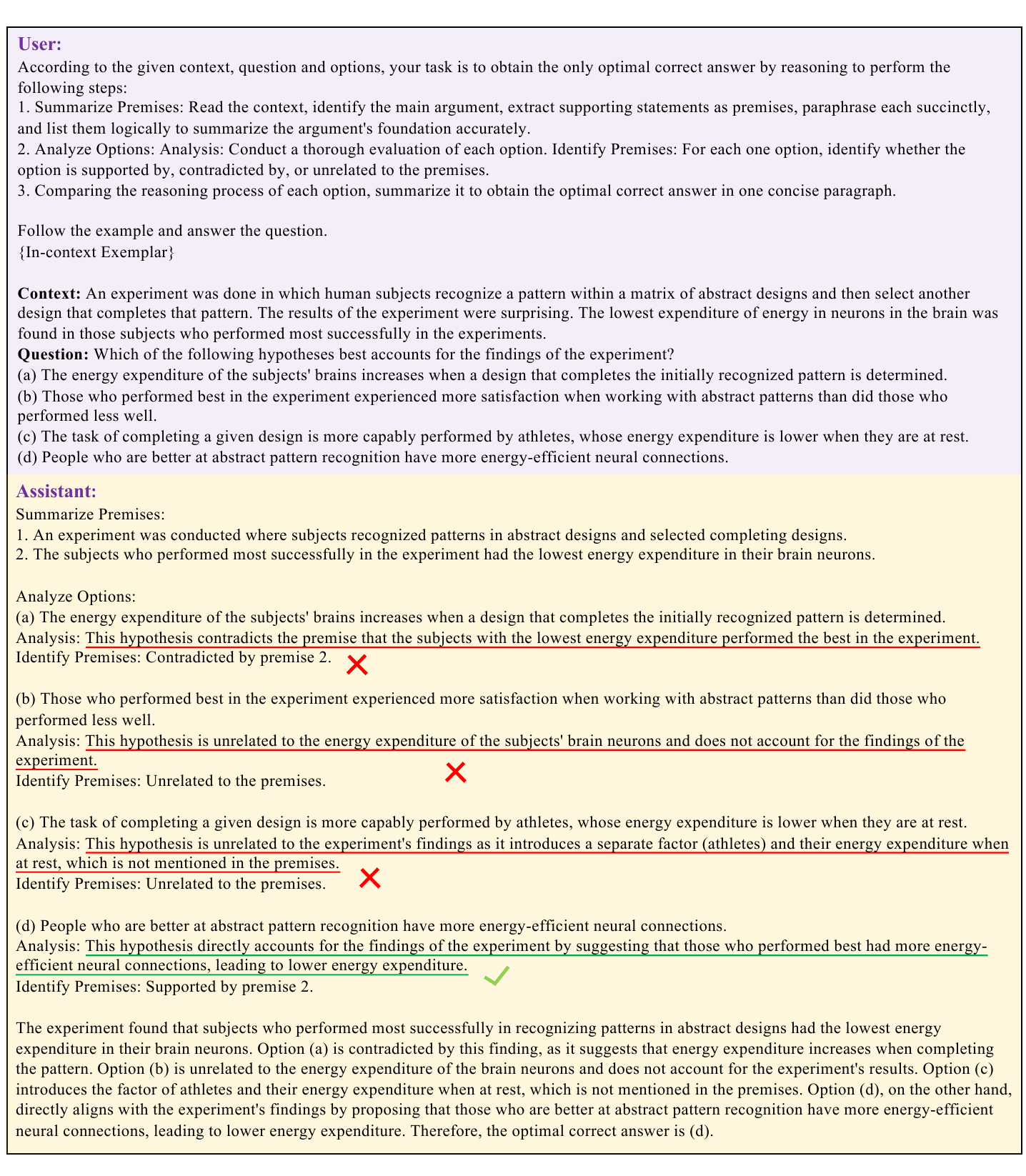}
    \caption{Example of the prompt used to annotate the CoT rationale, which involves the analyses for both correct and wrong options. The rationale also integrates summarizing premises and identifying premises for each option, forming the foundation for constructing new counterfactual instances.}
    \label{fig:cra}
\end{figure*}

\begin{figure*}[htb]

    \centering
    \includegraphics[width=2.1\columnwidth]{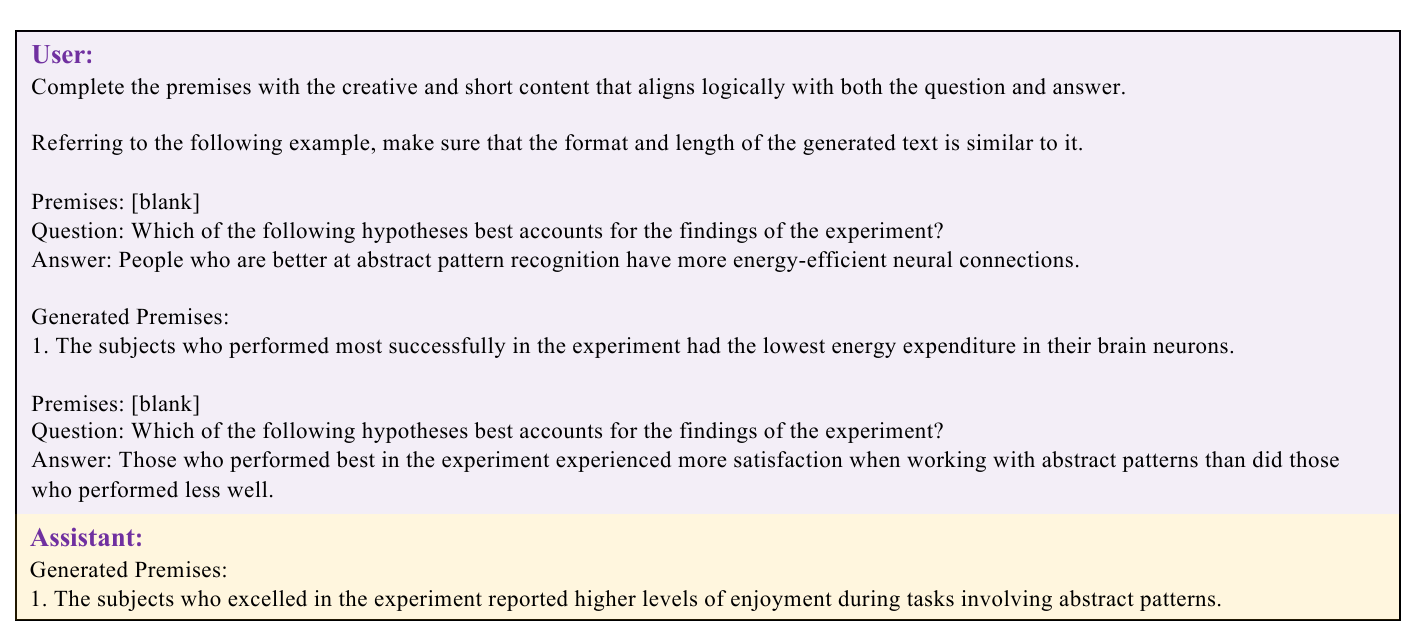}
    \caption{Example of the prompt used to generate counterfactual premises.}
    \label{fig:pg}
\end{figure*}

\begin{figure*}[htb]

    \centering
    \includegraphics[width=2.1\columnwidth]{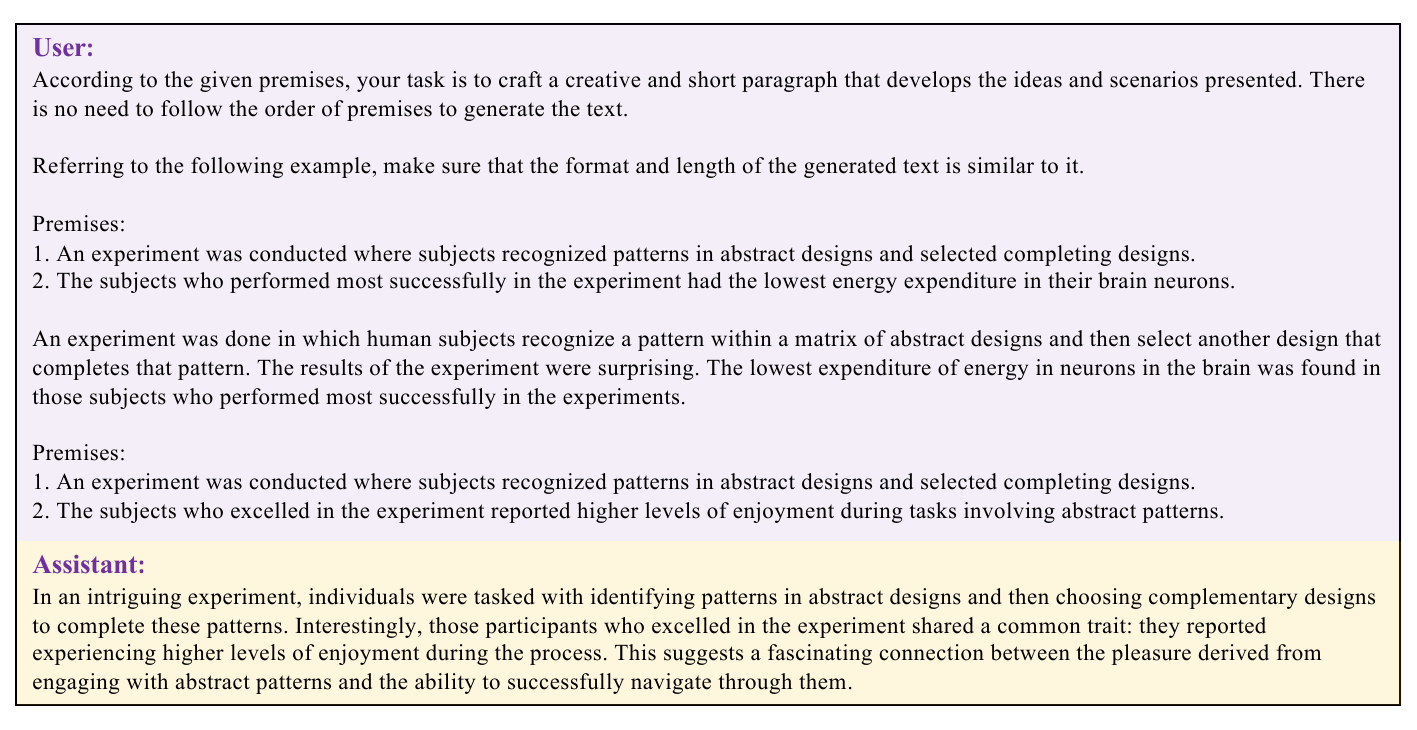}
    \caption{Example of the prompt used to generate counterfactual context.}
    \label{fig:cg}
\end{figure*}

\begin{figure*}[htb]

    \centering
    \includegraphics[width=2.1\columnwidth]{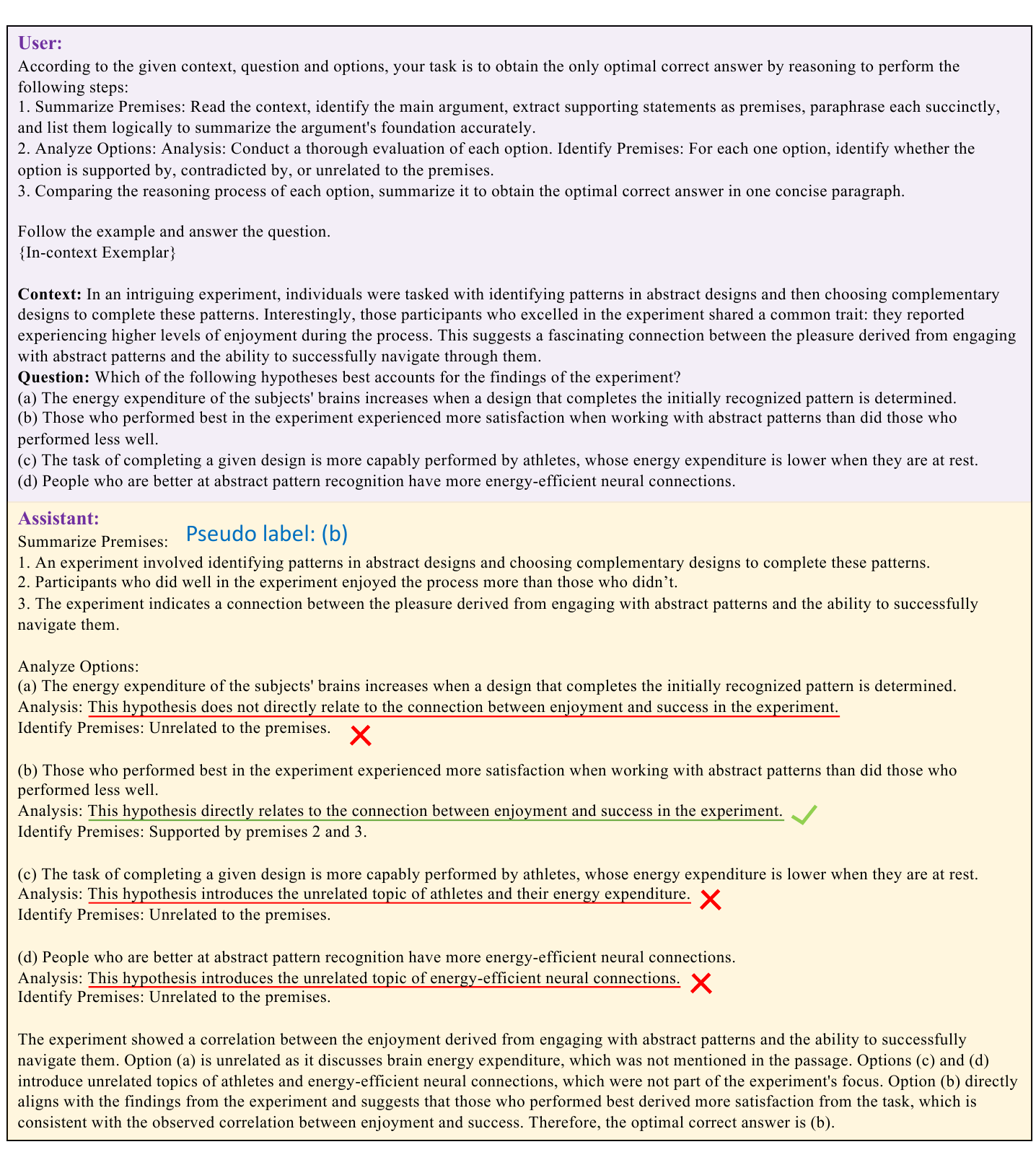}
    \caption{Example of the prompt used to validate whether the correct answer obtained through reasoning is consistent with the pseudo-label. Summarizing and identifying premises for each option are optional, and we maintain these intermediate steps to ensure that the rationale format remains consistent for both original and counterfactual samples.}
    \label{fig:cv}
\end{figure*}

\begin{figure*}[htb]

    \centering
    \includegraphics[width=2.1\columnwidth]{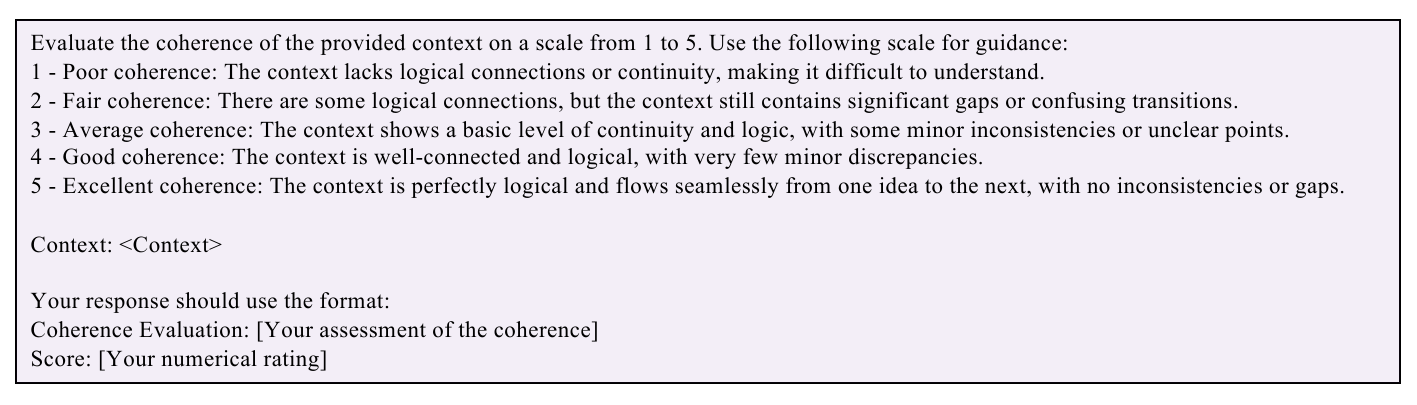}
    \caption{The prompt for evaluating the coherence of the context.}
    \label{fig:context_co}
\end{figure*}
\begin{figure*}[htb]

    \centering
    \includegraphics[width=2.1\columnwidth]{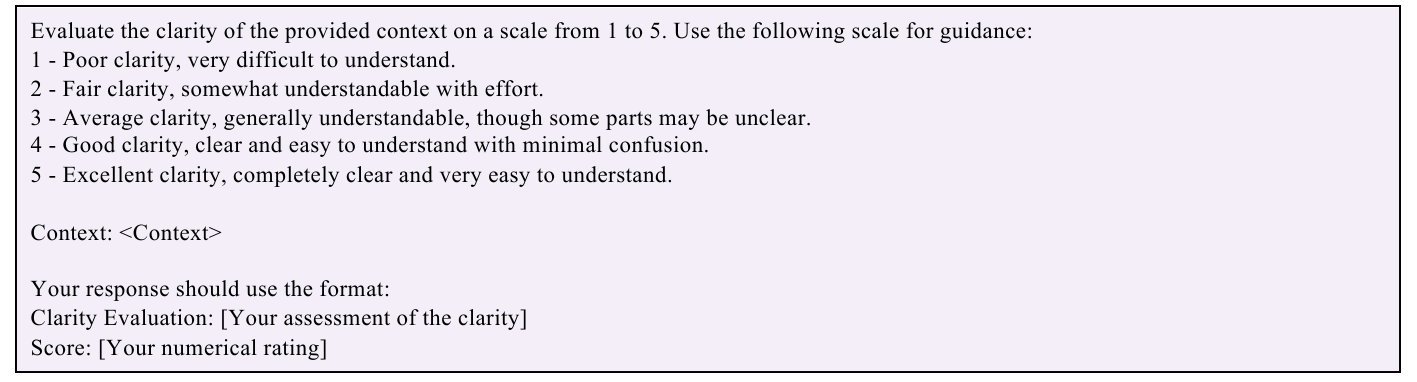}
    \caption{The prompt for evaluating the clarity of the context.}
    \label{fig:context_cl}
\end{figure*}

\begin{figure*}[htb]

    \centering
    \includegraphics[width=2.1\columnwidth]{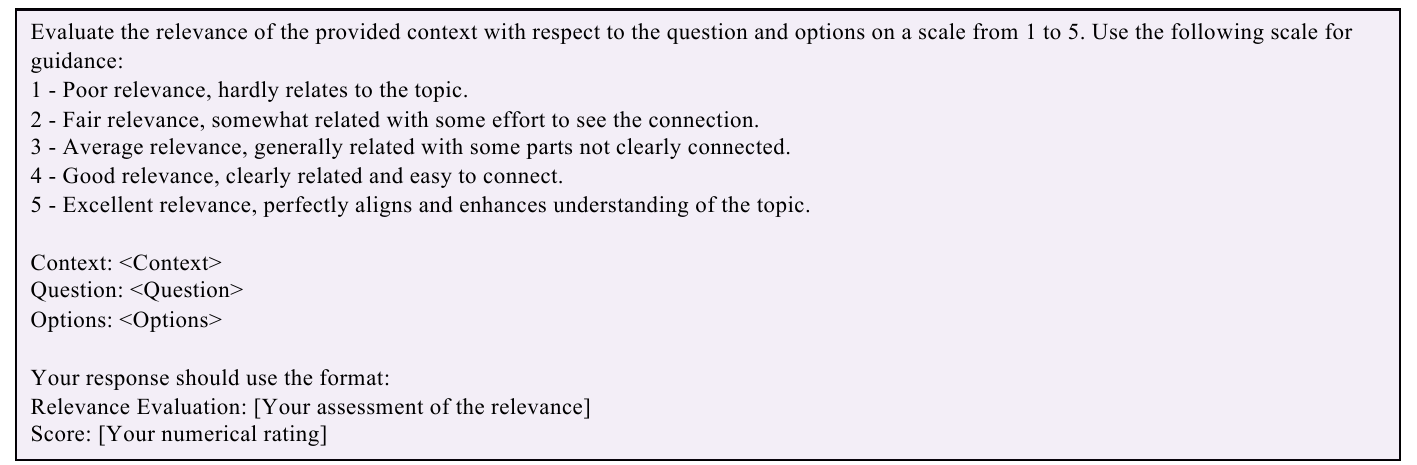}
    \caption{The prompt for evaluating the relevance of the context with respect to the question and options.}
    \label{fig:context_re}
\end{figure*}

\begin{figure*}[htb]

    \centering
    \includegraphics[width=2.1\columnwidth]{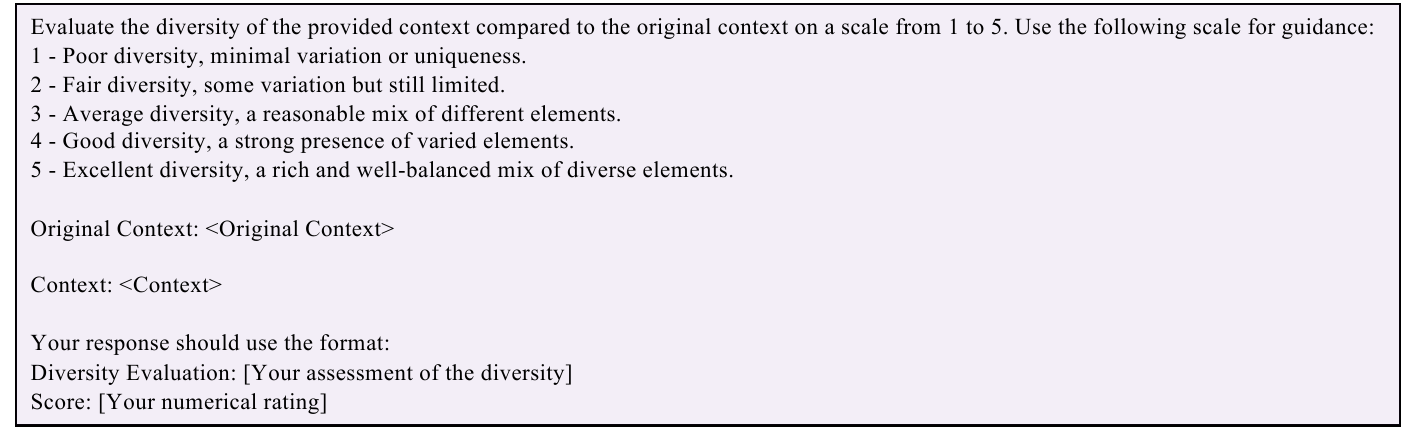}
    \caption{The prompt for evaluating the diversity of the context.}
    \label{fig:context_div}
\end{figure*}

\begin{figure*}[htb]

    \centering
    \includegraphics[width=2.1\columnwidth]{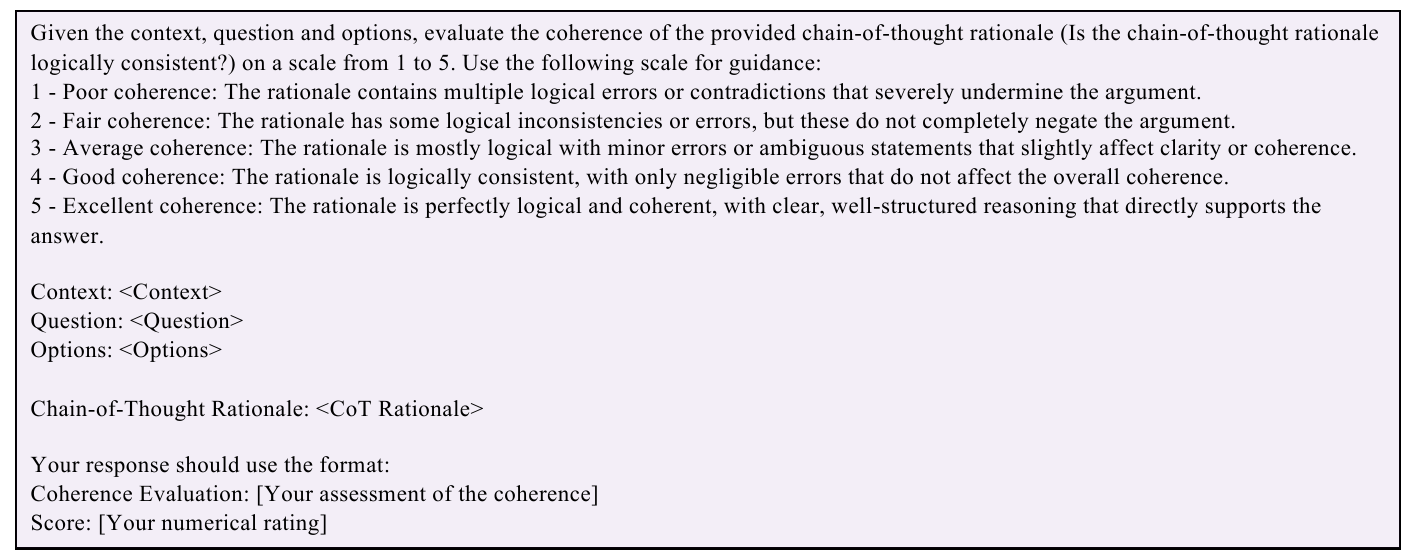}
    \caption{The prompt for evaluating the coherence of the CoT rationale.}
    \label{fig:cot_coherence}
\end{figure*}

\begin{figure*}[htb]

    \centering
    \includegraphics[width=2.1\columnwidth]{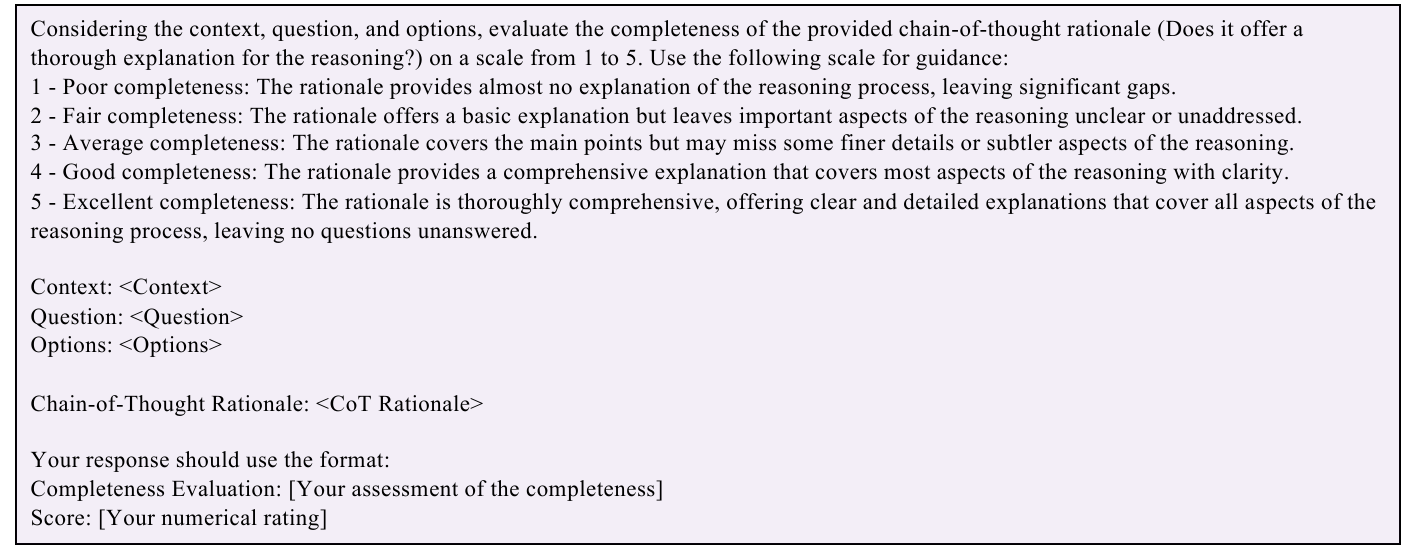}
    \caption{The prompt for evaluating the completeness of the CoT rationale.}
    \label{fig:cot_completeness}
\end{figure*}

\begin{figure*}[htb]

    \centering
    \includegraphics[width=2.1\columnwidth]{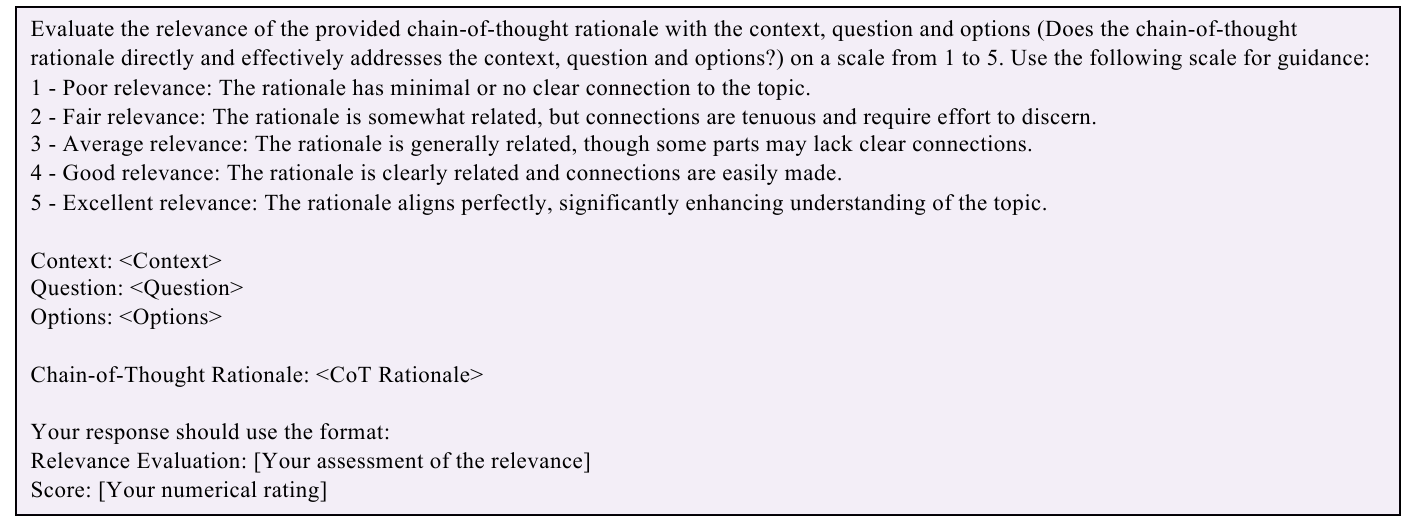}
    \caption{The prompt for evaluating the coherence of the CoT rationale with the context, question and options.}
    \label{fig:cot_relevance}
\end{figure*}

\begin{figure*}[htb]

    \centering
    \includegraphics[width=2.1\columnwidth]{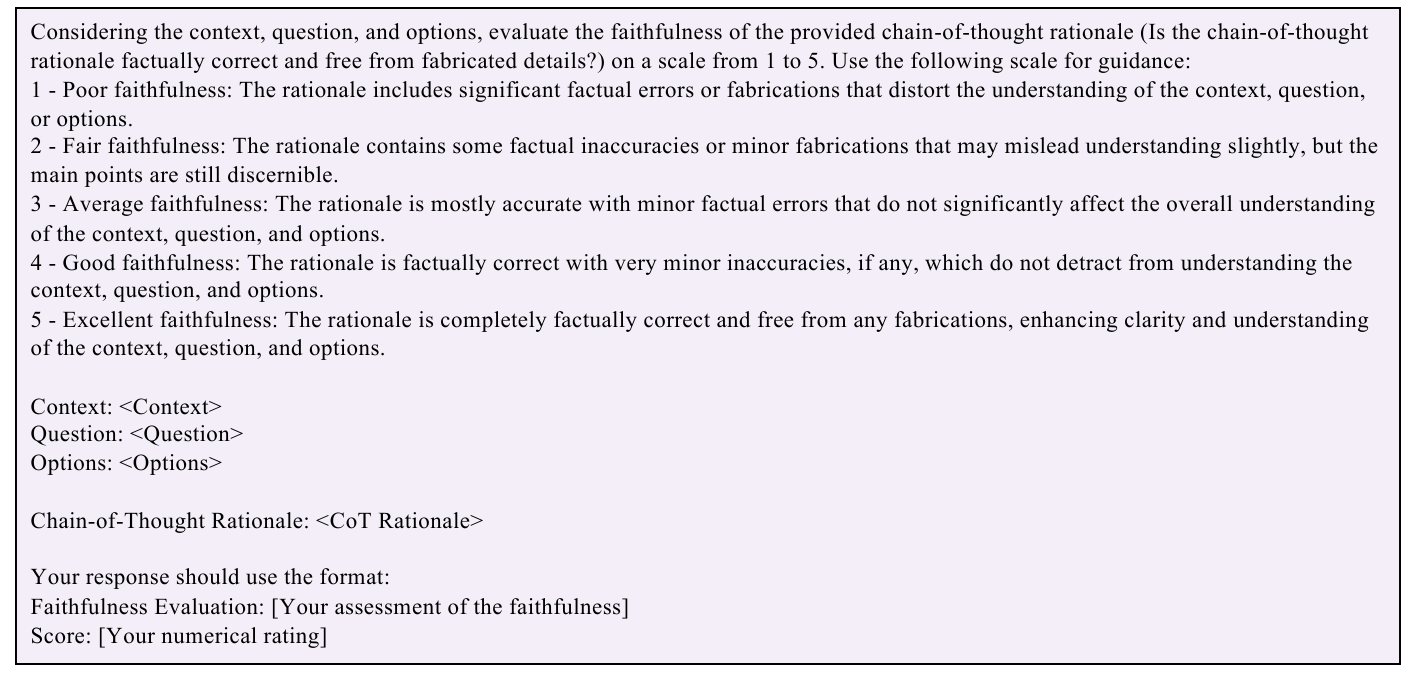}
    \caption{The prompt for evaluating the faithfulness of the CoT rationale.}
    \label{fig:cot_faithfulness}
\end{figure*}

\end{document}